\documentclass{article}

\usepackage[utf8]{inputenc} 
\usepackage[T1]{fontenc}    
\usepackage[colorlinks=true, allcolors=blue]{hyperref}
\usepackage{url}            
\usepackage{booktabs}       
\usepackage{amsfonts}       
\usepackage{nicefrac}       
\usepackage{microtype}      
\usepackage{graphicx}
\usepackage{natbib}
\usepackage{doi}
\usepackage{fancyhdr}
\usepackage{algorithmicx}
\usepackage[dvipsnames, table]{xcolor}
\usepackage{caption}
\usepackage{subcaption}
\usepackage{spverbatim}
\usepackage{float}
\usepackage{mathtools}
\usepackage{todonotes}
\usepackage{multirow}
\usepackage{amsmath}
\usepackage{amssymb}
\usepackage{soul}
\usepackage{nicematrix}
\usepackage{enumitem}
\usepackage{colortbl}

\usepackage{eqparbox,array}
\renewcommand\algorithmiccomment[1]{%
  \hfill\#\ \eqparbox{COMMENT}{#1}%
}

\newcommand{\Hdg}[1]{\textbf{#1}  } 

\definecolor{Boxgrey}{rgb}{1,1,1}
\definecolor{Outcol}{rgb}{.8, .2, .34}
\definecolor{Incol}{rgb}{.3, .5, .1}
\newcommand{\Prodz}{\rightarrow}
\newcommand{\Prod}{\;\;\rightarrow\;\;}
\newcommand{\LProdz}{\rightsquigarrow}
\newcommand{\LProd}{\;\;\rightsquigarrow\;\;}
\newcommand{\LLMin}[1]{\textcolor{Incol}{\ensuremath{#1}}}
\newcommand{\LLMout}[1]{\textcolor{Outcol}{\ensuremath{#1}}}

\usepackage{hyperref}

\usepackage[accepted]{icml2024}

\icmltitlerunning{SimLM: Can Language Models Infer Parameters of Physical Systems? A Technical Report}

\newcommand{\MethodName}{SimLM }

\begin{document}

\twocolumn[

\icmltitle{
SimLM: Can Language Models Infer Parameters of Physical Systems?
}



\icmlsetsymbol{equal}{*}

\begin{icmlauthorlist}
\icmlauthor{Sean Memery}{yyy}
\icmlauthor{Mirella Lapata}{yyy}
\icmlauthor{Kartic Subr}{yyy}
\end{icmlauthorlist}

\icmlaffiliation{yyy}{Department of Informatics, the University of Edinburgh, United Kingdom}

\icmlcorrespondingauthor{Sean Memery}{s.memery@ed.ac.uk}

\icmlkeywords{Large Language Models, Machine Prediction, Physics Simulation}
\vskip 0.3in
]



\printAffiliationsAndNotice{}  

\setlength{\abovedisplayskip}{1.5pt}
\setlength{\belowdisplayskip}{1.5pt}

\date{}

\begin{figure*}
\includegraphics[width=0.82\linewidth]{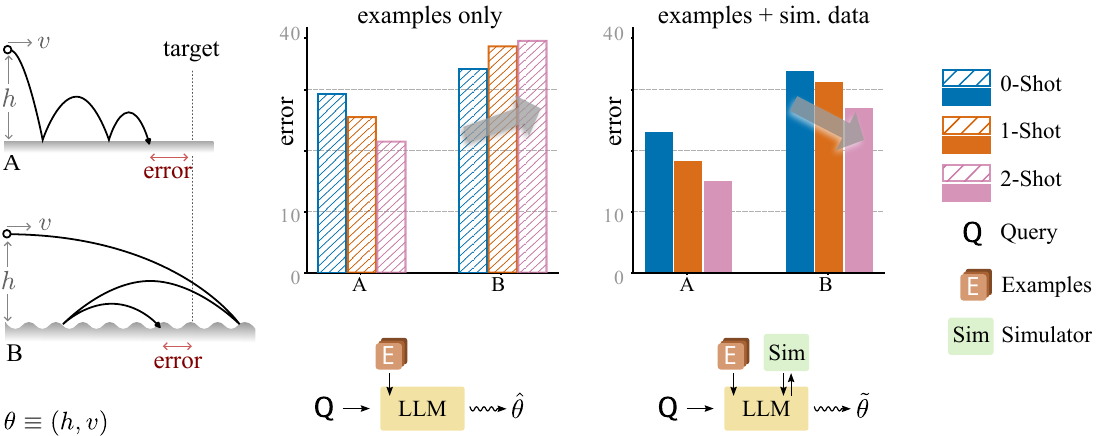}
\centering
\caption{We posed a number of Large Language Models a simple query $Q$: ``With what horizontal velocity $v$ and from what height $h$ should a ball be thrown so that its third bounce is within $1$ m of $50$ m''. Most models are incapable of answering this query. The figure shows the mean errors when the ground is flat (A) as well as sinusoidal (B). Surprisingly, providing more examples worsens the performance when the forward problem is difficult (sinusoidal case). The central proposition in  this article is that augmentation of queries/prompts via a physics simulator enhances the ability of LLMs to reason about physics for difficult problems.
}
\label{fig:main}
\end{figure*}

\begin{abstract}
Several machine learning methods aim to learn or reason about complex physical systems. A common first-step towards reasoning is to infer system parameters from observations of its behavior. In this paper, we investigate the performance of Large Language Models (LLMs) at performing parameter inference in the context of physical systems. Our experiments suggest that they are not inherently suited to this task, even for simple systems. We propose a promising direction of exploration, which involves the use of physical simulators to augment the context of LLMs. We assess and compare the performance of different LLMs on a simple example with and without access to physical simulation.

\end{abstract}

\section{Introduction}
The rapid evolution of language models has unveiled a powerful form of artificial intelligence that can generate plausible responses to general-purpose queries. These models have been found lacking in some domains and in this paper we demonstrate that they are fundamentally unsuited to reasoning problems involving physical systems. However, due to their sophisticated and articulated interpretation of complex ideas in natural language the hope is that, someday, they will be able to provide useful insights by answering questions such as ``Is `this' protein sequence likely to fold into a similar structure to `that' sequence?''. 

Indeed, we begin with a far simpler problem and assess the efficacy of LLMs at reasoning about physics--the \textit{inverse} problem. Specifically, we consider predicting the location of the third bounce of a spherical projectile launched from a certain height $h$ with a certain horizontal initial velocity $v$, for different terrain (see Figure~\ref{fig:main}). This is in contrast to concurrent works that investigate their ability to \textit{predict} behavior (the forward problem). 

Many methods~\citep{DBLP,Kojima2022LargeLM} improve reasoning abilities of LLMs via augmentations to their contexts. There have also been methods that rely on external systems to improve the consistency of LLM logic ~\citep{liuMindEyeGrounded2022,Madaan2022LanguageMO,Lyu2023FaithfulCR}. These methods improve capabilities while being independent of the training of the LLM. We follow this paradigm, but focus on physical reasoning. Rather than evaluate a model's ability to methodically dissect a physics problem into a sequence of solvable problems, as a first step we consider their ability to solve inverse physics problems. Specifically, we assess the abilities of LLMs to \textit{infer initial parameters} $\theta \equiv (h,v)$ in the above example of projectile motion. 

Several methods in machine learning aim to mimic human intuition about complex physical systems \citep{Spelke1992OriginsOK,Bramley2018IntuitiveEI} and develop approximate inference methods \citep{Lake2016WhatCT,Tenenbaum2018BuildingMT,Ullman2018LearningPP} for downstream applications such as \textit{system identification} in robotics \citep{LopezGuevara2017AdaptablePT}. There is some evidence that LLMs require `coaxing' to elicit reasoning natural to humans \citep{DBLP}. Few-shot chain-of-thought (CoT), and more sophisticated methods \citep{yaoReActSynergizingReasoning2023}, are promising directions for natural language, but are they effective for problems involving physics? 

In this paper we devise a simple inverse physics problem whose difficulty can be parameterised. Then we compare and analyse the performance of state-of-the-art LLMs at solving this problem. All chosen models perform unsatisfactorily. We find that there is benefit from using chain-of-thought reasoning to augment the context with examples of the forward problem (labeled `examples only' in Figure~\ref{fig:main}). For a flat surface, task A, the results show that few-shot examples consistently improve performance. However changing the surface to a \verb|sin| wave, task B, adds sufficient complexity to yield diminishing returns. 

We propose to ameliorate this by enabling the LLM to query a physical simulator. This is in the spirit of Retrieval-Augmented Generation (RAG) \citep{DBLP:journals/corr/abs-2005-11401}, where reference information is  retrieved using an external system. In our case, the external system is a physical simulator. Some improvement is obtained via this strategy which we call SimLM (pronounced sim-lim). Despite some success in the simple example, we provide further experiments in Section~\ref{sec:futurework} to draw the attention of the community to this general problem which remains open and exciting.


\section{Background}

\subsection{Previous Work}

\Hdg{Intuitive physics} There has been some work on the ability of humans to reason about some complex physical systems intuitively without formal training in physics. For example, on the behavior and learning patterns of infants \citep{Spelke1992OriginsOK,doi:10.1111/1467-8721.ep10770614,10.1093/acprof:oso/9780198524021.003.0004} and their ability to learn efficiently from sparse experiences. There are some models of how intuition is built \citep{Battaglia2013SimulationAA,Lake2015HumanlevelCL,Battaglia2016InteractionNF} surrounding physical events. Recently, generative AI and LLMs \citep{Wu2015GalileoPP,duan2022survey,Benchekroun2023WorldSenseAS} have been used to explore similar ideas. \cite{Jassim2023GRASPAN} investigate intuitive physical understanding of multi-modal models (language and image understanding) on a data set created from physical simulations towards labeling videos as depicting plausible physics events. 

\Hdg{Prompting and Reasoning}
There is considerable interest in devising augmentation and manipulation of the context provided to large language models so that its performance improves for a given task. This general method is known as ``prompt engineering'' \citep{Liu2021PretrainPA}. Several papers \citep{nye2021work,Kojima2022LargeLM} demonstrate that when an LLM is constrained to solve a task ``step by step'', it improves reasoning capabilities by triggering a favorable response for the LLM. When applied to calculating answers to maths word problems, such a dissection of the problem can make the difference between the LLM producing one incorrect answer to instead answering some sub-problems correctly. This is particularly known to benefit sufficiently large LLMs on tasks involving multi-step computations. With this idea, LLMs have been shown to perform many tasks zero-shot (i.e. no examples given) with appropriate prompting and augmentation. Many approaches have enhanced this fundamental idea recently \citep{Dua2022SuccessivePF,Creswell2022SelectionInferenceEL,Press2022MeasuringAN}.
%
\cite{Qiao2022ReasoningWL} provide a thorough survey of the LLM reasoning landscape that catalogs many methods. In particular, it lists many examples of single-step, strategy-enhanced reasoning, i.e. methods that improve reasoning using just a single step addition to the LLM context \citep{Kojima2022LargeLM,zhang2022automatic,Fu2022ComplexityBasedPF,Zhou2022TeachingAR}. Recently, new directions of LLM reasoning have been introduced that attempt to combine reasoning and action. \cite{yaoReActSynergizingReasoning2023} propose a system for iterative reasoning and action with impressive results involving grounded text-based plans within simulated text environments. \cite{haoReasoningLanguageModel2023} propose a Monte Carlo tree search algorithm to optimize reasoning towards grounded understanding of world models. 

\Hdg{LLMs and Physics}
Despite many works that evaluate LLMs on physics data \citep{XuanQuy2023EvaluationOC,Ding2023UsingLL,Yeadon2023TheIO,alidib2023physics}, only a handful of them explore the interactions of LLMs with physics environments. For example, \cite{liuMindEyeGrounded2022} introduces a system to ground LLM reasoning using physics simulation to answer predetermined physics-based questions. They use a secondary LLM step where the conditions of a physics simulation, defined explicitly in code, and statistics of the outcome are provided in-context to the LLM answering the original question. This yields a significant improvement over standard few-shot methods on physics QA data. However, it does not perform tasks within a simulated environment. \cite{Benchekroun2023WorldSenseAS} introduce a dataset to evaluate the consistency of LLM's world models, through repeated queries about specific physical situations. A recent benchmark for extensive physical reasoning, called NEWTON \cite{wang2023newton}, evaluates the ability of LLMs to predict physical properties of objects. Their method and data highlight the inability of LLMs to predict (forward model) the behavior of physical systems. In this paper we investigate how to improve the ability of LLMs to infer initial conditions of a physical system.

\subsection{Physical System}
The physical system in this paper is simple spherical projectile motion coupled with linear elastic collision at each bounce on the ground, as studied in secondary education level physics. The ground is flat for the simple experiment and bumpy for other experiments. The height $h_0$ and horizontal velocity $\vec{v}_0$ are initial conditions and acceleration due to gravity is $g$. We assume that the launch-point has an X-coordinate of zero. There may be other forces to consider in a realistic system, such as friction or air resistance but we do not consider them in this paper. 

The position $\vec{x}_t$ of the ball at any given time (\( t \)) is given by 
$\vec{x}_t = \vec{v}_0\;t + \vec{g}\;t^2/2$. 
Since the total mechanical energy (kinetic + potential) of an object remains constant when no non-conservative forces are acting upon it (for example air resistance, which is not considered in this work) we have $mgh_t + mv_t^2/2$ as a constant, where $m$ is the mass of the object.
Finally, when an object collides with the ground, the direction of its new velocity is governed by $\vec{n}$ the vector normal to the ground at the point of contact. The magnitude of the new velocity depends on the elasticity of the collision quantified by the coefficient of restitution $e$. For a perfectly elastic collision, $e = 1$, and for a perfectly inelastic collision, $e = 0$. In this paper, we use $e=0.9$ for all our experiments. The velocity after collision $\vec{v'}$ is given by $\vec{v'} = e\;\vec{r}$, where $\vec{r}$ is found by reflecting the incident velocity vector about the normal vector, $\vec{r} = \vec{v} - 2\;\vec{n}\;(\vec{v}\cdot\vec{n})$. 



\subsection{Prompting Methods Notation}
We extend existing notation \citep{sumers2023cognitive} that represents LLM prompting methods using production systems. 
For example, $X \Prodz X\LLMin{Y}\ $ represents a precondition (input) $X$ that results in an output of $XY$, where the production system specifies the possible values of $Y$. This was expanded to include large language models as they can be seen as providing probabilities across productions from which a final production is sampled. This was written as a stochastic production $Q \LProdz Q\LLMout{A}\ $ where $Q$ is the query and $A$ is the LLM generated answer. We adopt this notation as a tool to understand and compare recent prompting methods for LLMs as well as to denote our contribution. We use different colors to highlight context augmentation inputs (e.g. \LLMin{O}) vs LLM-generated outputs (e.g. \LLMout{A}). 

\paragraph{Retrieval-Augmented Generation}
As LLMs are inherently stateless, retrieval is a method commonly used to provide a working memory for the model. It can also be used to retrieve specific information that the LLM may use to improve on its completion. For example, a simple retrieval of relevant Wikipedia pages would improve an LLM's question-answering capability. This method is known as Retrieval-Augmented Generation (RAG) \citep{DBLP:journals/corr/abs-2005-11401} and can be denoted 
as
$
Q \Prodz Q\LLMin{O} \LProdz QO \LLMout{A} 
$
where $O$ is the retrieved information.

\paragraph{Self-Critique}
\cite{saunders2022selfcritiquing} introduced self-critique as a simple method that asks the LLM to critique its own output. For question-answering this is as simple as the LLM answering the question, critiquing its own answer, and answering again. Using the same notation as above, this can be written as a sequence of stochastic productions
\begin{equation}
Q \LProd Q\LLMout{A_0} \LProd QA_0\LLMout{C} \LProd QA_0C\LLMout{A_1} 
\end{equation}
where $C$ is the self-critique generated by the LLM based on its own initial answer \LLMout{A_0} which is then updated to \LLMout{A_1}.

\paragraph{Few-shot Chain-of-Thought}
\cite{DBLP} introduced chain-of-thought (CoT) prompting, a method to improve the reasoning abilities of LLMs. CoT prompting provides the LLM with in-context examples of question-reasoning-answer sequences. More of these examples has been shown to lead to better performance in general. This is visualized below, with each example $E$ containing full chain-of-thought:
\begin{equation}
Q \LProd \LLMin{E_1... E_N}Q\LLMout{R} \LProd E_1... E_NQR\LLMout{A}
\end{equation}

\section{\MethodName}
We now describe our method \MethodName of using a physical simulator in conjunction with existing work on prompting.

\subsection{Method}\label{sec:method}
Our prompting method inserts physics simulations between reasoning and critiquing steps. It can be viewed as a retrieval augmented generation combined with self-critique $C$ of its reasoning--- here the ``retrieve'' operation corresponds to physics simulation feedback $P$. Thus we can write this as a sequence of four productions where the second production is the execution of a physics simulator:
\begin{eqnarray}
    \nonumber Q & \LProd & Q\LLMout{R} \\
    \nonumber & \Prod & QR\LLMin{P} \\
    \nonumber & \LProd & QRP\LLMout{C} \\
    & \LProd & QRPC\LLMout{A} 
\end{eqnarray}
We apply this iteratively $N$ times, so that the model is allowed to improve on its reasoning  via multiple physics simulations and self-critique steps i.e. generate reasoning $R$, augment with feedback $P$, generate self-critique $C$ and repeat $N$ times before producing the final answer $A$:
\begin{eqnarray}
\label{eq:zeroshot}
    \nonumber Q & \LProd & Q\LLMout{R} \\ 
    \nonumber & \LProd & QR\LLMin{P_1}\LLMout{C_1} \\
    \nonumber & \LProd & QRP_1 C_1 ... \LLMin{P_n}\LLMout{C_n} \\
    & \LProd & QRP_1 C_1 ... {P_n}{C_n} \LLMout{A}
\end{eqnarray}
We refer to this method as `zero-shot with physics'. 

Providing in-context examples \citep{NEURIPS2020_1457c0d6}  to the LLM improves performance on many tasks. Chain-of-thought prompting showed this to be true for reasoning. Inspired by these findings, we extend our zero-shot method (Equation~\ref{eq:zeroshot}) to include examples $E_i$, $i=1,2,3\dots,m$. Assembling these elements, our full method is: 
\begin{eqnarray}
    \nonumber Q  & \Prod & \LLMin{E_1 ... E_m}Q \\
    \nonumber & \LProd &  E_1 ... E_mQ\LLMout{R} \\ 
    \nonumber & \LProd & E_1  ...  E_m QR\LLMin{P_1}\LLMout{C_1} \\ 
    \nonumber & \LProd & E_1  ...  E_m QRP_1 C_1 ... \LLMin{P_n}\LLMout{C_n}  \\
    & \LProd & E_1  ...  E_m QRP_1 C_1 ... {P_n}{C_n} \LLMout{A} 
\end{eqnarray}
This can also be summarized succinctly as 
\begin{equation}
Q \LProdz \{E\}^* QR \LProdz \{E\}^* QR\{PC\}^+ A
\end{equation} 
where we use $^*$ and $^+$ as used in regular expressions to mean any number of instances and at least one instance respectively.

\subsection{Implementation}\label{sec:implementation}
\paragraph{Physics Simulation}
To perform the physics simulation, we use the \verb|pymunk|\footnote{\url{https://github.com/viblo/pymunk}} package. This is a lightweight physics package for \verb|Python| that is built on top of the \verb|Chipmunk|\footnote{\url{https://chipmunk-physics.net/}} 2D physics simulator. \verb|Pymunk| performs rigid-body physics simulation with high efficiency and accuracy. To reduce errors caused by the physical limitations of the simulation, the frames per seconds of the simulation is set to 1000Hz. Doing so limits precision errors during physics calculation causing the simulation to be as deterministic as possible. 

Using \verb|Pymunk|, any arbitrary function $f(x)$ can be specified as the ground geometry by defining $x$ coordinate limits, e.g. $[-1000, 1000]$, and iterating between these limits with some step $c$. Then, a list of $y$ coordinates can be calculated as $y = f(x)$, and used with the $x$ positions to create a piece-wise linear approximation to the function. 

\paragraph{LLM Inference}
We perform LLM inference using the \verb|guidance|\footnote{\url{https://github.com/guidance-ai/guidance}} library. \verb|guidance| allows for consistent, template based prompting of multiple models, local and API-based. Using this, we created a pipeline to prompt LLMs, run simulations, show simulation results to those LLMs, and continue prompting. Algorithm \ref{alg:prompting} gives an overview of this pipeline. See the supplementary information for more detailed examples and prompts.

\begin{algorithm}
\caption{LLM prompting in SimLM}\label{alg:prompting}
$Ex = \verb|retrieve|(Q)$ 
\newline
$R, A = \verb|LLM|(Ex, Q)$ 
\newline
$P_1 = \verb|sim|(A)$
\newline
$C_1, A_1 = \verb|LLM|(Ex, Q, R, P_1)$ 
\newline
$P_2 = \verb|sim|(A_1)$ 
\newline
\dots 
\newline
$C_N, A_N = \verb|LLM|(Ex, Q, \dots, P_N)$  
\end{algorithm}

An important aspect of our method is the number of attempts of prediction and self-critique. In our implementation, we prompt the LLM to finish repeating attempts if it finds that it is meeting the requirements of the task. We found a suitable limit to be $N=5$ and use it in all experiments.

Many works have shown the importance of supplying LLMs with high quality examples for the target task. In our method, the LLM is given examples to act as long-term experience gain by the model. We use previous successful runs of an experiment as examples for future runs. In each experiment we have a success criteria, if this is met then the entire context is saved in a database of reasoning examples. In order to limit the context size of future trials, we then perform basic summarisation on the saved examples. See the supplementary information for summarisation details. Then, a database of summarized examples can be sampled from to provide future trials with in-context CoT examples. In order to prevent performance leakage, we limit the retrieval to examples created by \textit{the same model} on \textit{the same experiment}.

\begin{figure*}
    \centering
    \includegraphics[width=0.85\textwidth]{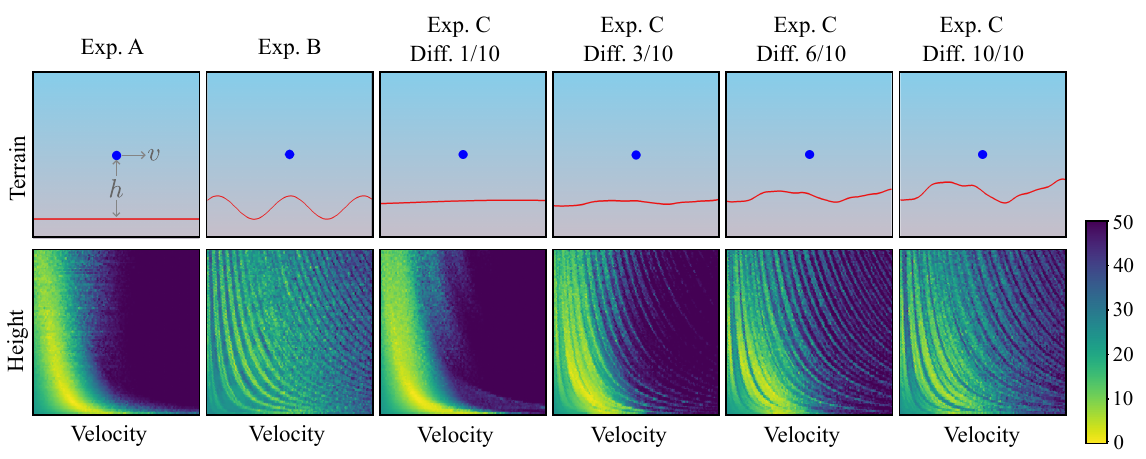}
    \caption{Visualizations capturing the difficulty of problems as a function of the ground surface (top row) and mean error across trials (bottom row). Errors (heat map colors) are shown across the parameter space spanned by  horizontal velocity (X-axis) and initial height (Y-axis) measured as the mean distance away from the target (50m). 
    }
    \label{heatmaps}
\end{figure*}

\section{Experiments}
We conducted three experiments (A, B, and C) where LLMs are asked to predict the initial conditions of a ball, such that its third bounce will occur within $1$m of a target distance of $50$m. The target and tolerance is fixed across all experiments for consistency. Across the variants we vary only the shape of the terrain the ball bounces upon. The top row in Figure~\ref{heatmaps} illustrates a few instances of this problem showing the ball (blue dot) and the terrain (red line).

The LLMs output two values, as a JSON object: \verb|height| and \verb|horizontal_velocity| which are used to initialize the ball in the simulation. Other initial conditions, such as starting $x$ position and vertical velocity, are set to zero. We compare our method (\MethodName) against a baseline Chain of Thought (CoT) method. The baseline simply uses the first set of initial conditions output by the model, with no physics simulation feedback. This ``blind'' attempt at reasoning about the initial conditions does not update itself based on simulation results. Our method uses feedback from simulation results and the LLM's self-critique.

We provide examples to the LLM that are gathered from previous runs of the same experiment. This is consistent across all experiments, but retrieved examples are not using the same target value as the current trial. We retrieve a random selection of examples that are sampled uniformly across targets, with buckets of 10m. Thus each trial is equally likely to contain examples that are near or far from its target (rather than biased examples around the target). 

\paragraph{Flat Surface}
Experiment A is a simple baseline of the performance of LLMs with and without our method. The surface that the ball is bouncing on is a flat plane described by $y = 0$. There are still however multiple unknown factors such as the coefficient of restitution of the ball and surface to consider. This makes it difficult for an LLM to explicitly calculate the required parameters, as we want to evaluate the LLM's common sense physical reasoning rather than its computation abilities. Column $1$ of figure \ref{heatmaps} shows a visualization of this experiment.

\paragraph{Uneven Surface}
Experiment B is considered a non-linear step from experiment A. On a flat surface, each bounce should be a further distance from the ball's origin. However in experiment two, the surface is described by the sinusoid $y = \alpha \cdot \sin( f \cdot x )$, where $\alpha$ is the amplitude of the sinusoid, and $f$ is the frequency. This can lead to the ball colliding with the surface at an incline and bouncing backwards, causing the ball's bounce positions to no longer be monotonic. For an LLM, this is a particularly difficult challenge as examples are not as helpful when a small difference in position can lead to extremely different outcomes. The LLM must then rely heavily on particularly relevant examples and trial-and-error to improve its prediction. We fix the amplitude and frequency of the sinusoid to $1$. Column $2$ of figure \ref{heatmaps} shows a visualization of this experiment.

\paragraph{Varying Difficulty}
In experiment C we explore how the performance of an LLM degrades as a surface becomes more ``difficult''-- that is more chaotic or rough. The parameter space heatmaps of figure \ref{heatmaps} illustrate the difficulty posed to an optimizer during inference across experiments. For different terrains, as the area in the heatmap where the error is lowest becomes sparser and less defined, we classify the terrain as becoming more ``difficult'' to predict. We generate a range of settings by interpolating between two surfaces, one considered easy and one considered difficult. The former is a single sinusoid
    $y_E = 0.15\cdot \sin(0.25x)$
while the latter is defined as a sum of sinusoids:
\begin{equation}
    \nonumber y_H = 0.6\cdot \sin(0.9x) + 0.15\cdot \sin(2.25x) + 0.05\cdot \sin(4.5x)
\end{equation}
Then, to create a surface with a difficulty rating of $d \in [0,1]$, a simple linear interpolation is performed between the two surface functions. We create $10$ different surfaces for experiment three, of uniformly spaced difficulty, in order to evaluate how baseline CoT and \MethodName adapt to increasingly difficult surfaces. Columns $3-6$ of figure \ref{heatmaps} visualize these surfaces, for difficulties $1/10$, $3/10$, $6/10$, and $10/10$.

\newcommand{\tabsp}{ \rule{0pt}{1\normalbaselineskip}}
\begin{table*}
    \renewcommand*{\arraystretch}{0.75}
    \centering
     \begin{NiceTabular}{rl|cc>{\columncolor[rgb]{1, 0.9, 0.9}}c|cc>{\columncolor[rgb]{0.9, 1, 0.9}}c}
        & & \multicolumn{3}{c}{Baseline CoT} & \multicolumn{3}{c}{\MethodName} \\
        \rule{0pt}{0.8\normalbaselineskip}
        & & \textit{0-Shot} & \textit{1-Shot} & \textit{2-Shot} & \textit{0-Shot} & \textit{1-Shot} & \textit{2-Shot} \\            
        \hline
        \tabsp
        \multirow{2}{*}{PaLM-2}            
        \RowStyle{\color{gray}}    
        & \textit{Exp. A} & 34.84 & 28.51 & 22.41          & 24.21 & 19.34 & 15.03\\
        & \textit{Exp. B} & 32.12 & 35.53 & 37.54          & 31.70 & 30.72 & 27.05 \\
        \tabsp
        \multirow{2}{*}{GPT-3.5-turbo}
        \RowStyle{\color{gray}}    
        & \textit{Exp. A} & 19.23 & 18.68 & 16.40         & 20.18 & 16.34 & 15.41 \\
        & \textit{Exp. B} & 37.03 & 40.39 & 40.19         & 42.17 & 36.20 & 29.10 \\
        \tabsp
        \multirow{2}{*}{Llama-2-70b-chat} 
        \RowStyle{\color{gray}}    
        & \textit{Exp. A} & 29.14 & 25.95 & 21.67 & 23.64 & 20.53 & 19.45 \\
        & \textit{Exp. B} & 40.91 & 41.62 & 40.32 & 40.12 & 35.54 & 30.15 \\
        \tabsp
        \multirow{2}{*}{Llama-2-13b-chat} 
        \RowStyle{\color{gray}}    
        & \textit{Exp. A} & 30.18 & 26.48 & 22.96          & 26.52 & 23.53 & 20.87 \\
        & \textit{Exp. B} & 43.89 & 44.68 & 43.63          & 42.19 & 39.67 & 31.16 \\
        \tabsp
        \multirow{2}{*}{Llama-2-7b-chat}  
        \RowStyle{\color{gray}}    
        & \textit{Exp. A} & 41.95 & 29.03 & 27.99          & 42.73 & 29.38 & 25.16 \\
        & \textit{Exp. B} & 43.02 & 41.09 & 44.02          & 42.10 & 34.87 & 31.30 \\
     \end{NiceTabular}
    \vspace{2mm}
    \caption{The average absolute error (for a target of 50m) of LLMs for experiments A and B, for baseline CoT and \MethodName methods.}
    \label{tab:combined}
\end{table*}

\section{Results}

We tabulate mean errors obtained using different models (Table \ref{tab:combined}) across Experiments A and B.
We calculate an average error, in each case, across $\sim$1000 instances. We plot
\footnote{The colors chosen for plots were based on \cite{pointsofview}, to maximisze readability for colour blindness.}
comparisons for the models PalM-2 (\verb|chat-bison| variant) and GPT-3.5-turbo (figure \ref{palm_vs_gpt}). 


The results of experiment 3 for the PaLM-2 model are presented in figure \ref{exp3:palm}. Here, surface difficulties are grouped into the categories \textit{easy} (difficulty 1-3), \textit{medium} (difficulty 4-7), and \textit{hard} (difficulty 8-10). For each category, the results of the corresponding surfaces are averaged and displayed as a relative error of \MethodName over baseline CoT. That is, a relative error value below $1$ signifies better performance of \MethodName and decreasing relative error means a growing disparity between the \MethodName results and baseline CoT. 

We only include average performances and selected statistics in the paper. Complete results of each experiment are included in the appendix. For each comparison between baseline and non-baseline results, two-sample t-tests were performed to assess whether there were statistically significant differences in their distributions. In each test, the null hypothesis, that there is no difference between the groups, was tested. A threshold p-value of $0.05$ was used to determine statistical significance. In each case the p-value was below this threshold, and the null hypothesis was rejected, suggesting that the differences observed in the results were unlikely to have occurred by chance.

\section{Discussion}

\subsection{Models Compared}
We focus on three classes of models as a selection of currently feasible models (as of December 2023), based on a balance of efficiency, cost, and capability:
\begin{itemize}[topsep=-.3em,itemsep=-.3em,leftmargin=1em,itemindent=.5em]
    \item \verb|GPT-3.5-turbo|: The API accessible chat model from OpenAI \citep{NEURIPS2020_1457c0d6}.
    \item \verb|PaLM-2|: The API accessible model from Google \citep{anil2023palm}, we used the \verb|chat-bison| variant.
    \item \verb|Llama-2|: Open source, local chat models from Meta \citep{touvron2023llama} with 7b, 13b, and 70b parameters.
\end{itemize}

\begin{figure}
    \centering
    \includegraphics[width=0.45\textwidth]{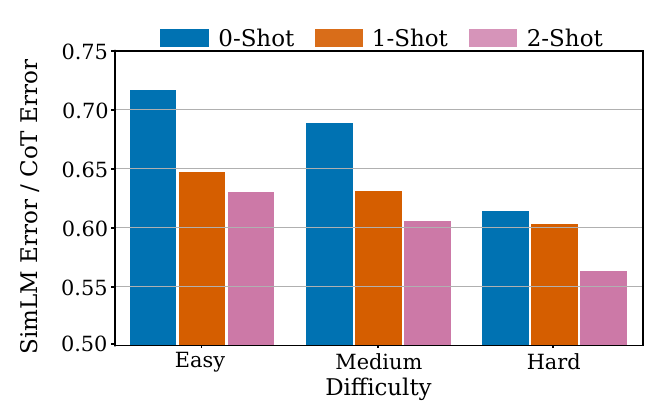}
    \caption{A plot of the error ratio of out method vs baseline for surfaces with varying difficulty. The error ratio is less than one showing that \MethodName outperforms the baseline. As the problem becomes more difficult, \MethodName is relatively more effective.}
    \label{exp3:palm}
\end{figure}


\subsection{Simulation Improves Performance}
\begin{figure}
    \centering
    \includegraphics[width=0.85\linewidth]{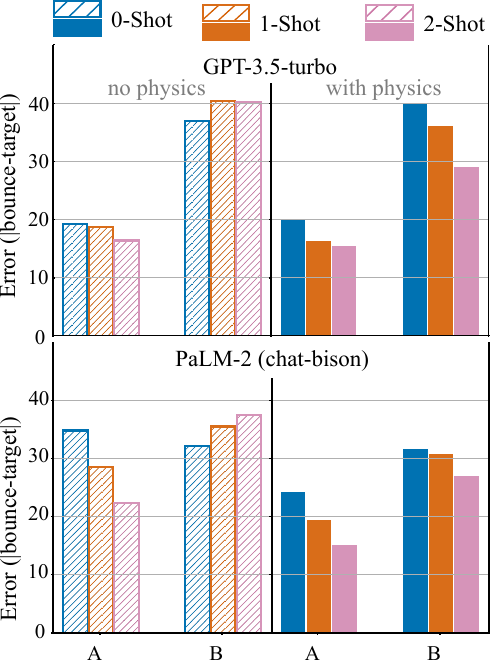}
    \caption{The use of physics simulation data helps LLMs (GPT and PaLM) cope with increasing difficulty. Plots of error in choosing initial conditions for a ball to bounce at a target distance on a flat surface (experiment A) and on an uneven surface (experiment B).}
    \label{palm_vs_gpt}
\end{figure}

In figure \ref{palm_vs_gpt}, we observe little difference between the LLM performance across methods in experiment A. However, notice the decrease in error between experiment B with no physics feedback (baseline CoT) and experiment B with physics feedback (\MethodName). We see a small drop in error for \MethodName over the baseline, but significantly the baseline CoT does not effectively make use of examples in experiment B, and performance actually suffers as more examples are used. On the other hand, \MethodName uses simulation to enable effective use of examples and improve performance. This is an important aspect of our method, to improve physical reasoning on uneven terrains.

\subsection{Relative Error Decreases as Difficulty Increases}
Figure \ref{exp3:palm} highlights the performance increase of \MethodName over the baseline for the best performing model, PaLM-2 (\verb|chat-bison| variant). As the difficulty of the surface increases the overall relative error decreases. \MethodName is outperforming baseline CoT more as the surface becomes more difficult and uneven. We contribute this to the LLM relying more on the feedback received from the physics simulation, as the surface becomes more difficult to predict. We also see that the relative error of the two methods is below $1$ in all cases, showing the strength of \MethodName on uneven surfaces generally, and that the relative error decreases as more examples are used in each method, meaning \MethodName is benefiting more from previous examples than baseline CoT. 

\subsection{Large Models Outperform Small}
As a further study of the capabilities of smaller models, in particular the popular range of Llama-2 models, we repeat experiments A and B for the Llama-2 models of parameter sizes 7b and 13b, and compare these results to the largest Llama-2 model of parameter size 70b. Table \ref{tab:combined} shows the results of the model size study.

We observe a similar trend that is seen in the larger, API based, models. Experiment A is still a relatively achievable task for the models, but as the difficulty is increased in experiment B the same loss of CoT improvements occur. The overall difference between models is simply a decreasing average error as the model size increases, as generally expected. This is a sign that larger models perform better at physical reasoning than their smaller counterparts, meaning there is an ability inherent in the models that is being improved on when parameter counts are scaled up.

\begin{figure}[htbp]
    \centering
    \includegraphics[width=0.45\textwidth]{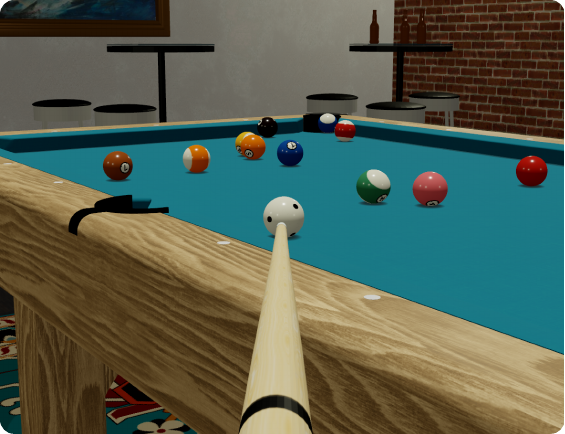}
    \caption{An LLM taking a shot using the \textit{PoolTool} simulator.    \label{pooltool}}
\end{figure}

\section{Generalisation and Future Work}\label{sec:futurework}
Although our proposed idea was useful in inferring two parameters of interest, we encountered difficulties in higher-dimensional settings. For example, it was not very successful on an experiment, related to the physics on a pool/billiards table, with five parameters. Pool has been used for research purposes \cite{Leckie2006AnEP, Smith2007PickPocketAC}. We explain the experiment and have set up an evaluation framework (which we will share) with leader-boards in the hope that evaluation of future work in this direction may be streamlined.

\begin{table}[htbp]
    \centering
    \begin{tabular}{c|c|c}
            \textit{(Random = 0.4\%)}      & CoT Baseline & \MethodName  \\
                  \hline
    GPT-3.5-turbo &   0.7\%  &   1.1\% \\
    PaLM-2    &   1.4\%  &   2.0\% \\
    Llama-2-13b-chat &   0.6\%  &   0.9\% \\
    Llama-2-7b-chat &   0.5\%  &   0.8\% \\
    \end{tabular}
    \caption{Success rate of LLMs potting balls in 3D pool.}
    \label{tab:experiment3D}
\end{table}

We posed LLMs with the query ``Pot the red ball into the back left pocket'' and measured their success rate (see Table~\ref{tab:experiment3D}).  In order to provide the LLM with a physically-accurate 3D environment, we made use of the \verb|PoolTool|\footnote{\url{https://github.com/ekiefl/pooltool}} application, a highly accurate simulation of the game of pool. \verb|PoolTool| is capable of simulating ball spin, sequences of ball collisions, cushion reflections, and many other physical interactions relevant to pool/billiards. In order to take a shot in \verb|PoolTool|, five parameters must be set:
\begin{enumerate}[topsep=-.3em,itemsep=-.3em,leftmargin=1em,itemindent=.5em]
    \item \verb|V|: The speed at which the cue ball will travel, $\verb|V| \sim [0,5]$.
    \item \verb|theta|: The angle of the cue, in order to impart different levels of spin, $\verb|theta| \sim [0,90]$.
    \item \verb|phi|: The azimuth of the shot, the angle corresponding to the direction the cue ball will travel in, $\verb|phi| \sim [0,360]$.
    \item \verb|a|: The east-to-west coordinate of the point of contact of the cue to the cue ball, $\verb|a| \sim [-1,1]$.
    \item \verb|b|: The north-to-south coordinate of the point of contact of the cue to the cue ball, $\verb|b| \sim [-1,1]$.
\end{enumerate}
Through the combination of these parameters, many different kinds of shots can be taken. For example, in order to apply backspin to the cue ball \verb|theta| should be set to a high value and/or \verb|b| should be set to a low value. 

We extended \verb|PoolTool| to include string representations of events in the simulation of each shot. A sequence of events can then be represented as a chain of sentence descriptions, for example:
\begin{itemize}[topsep=-.3em,itemsep=-.3em,leftmargin=1em,itemindent=.5em]
    \item \verb|CollisionEvent(b1, b2)| $\rightarrow$ \textit{``Ball b1 collided with ball b2 at position (x,y).''}
    \item \verb|PotEvent(b, p)| $\rightarrow$ \textit{``Ball b fell into pocket p.''}
\end{itemize}
We also include a textual description of the current board state, before and after each shot using a chain of \verb|"ball_id=(x,y)"| for each ball with position $(x,y)$ on the table. The positions of the pockets are also given. All of this information is included when prompting the LLM, as we did in our original 2D problem. However, the information needed to describe a single simulation run consists of a large body of text. In order to remain in the context limit of most models, the method was allowed at most $N=3$ retries at the task before resulting in failure if there is no success. We also fix the examples used for baseline and \MethodName to $3$, and retrieve previously successful attempts similarly to the 2D case. 

While a wider variety of tasks is supported, for this study we limit tasks to the most basic action in pool:\textit{ potting a single ball}. We task LLMs with potting a ball (placed randomly within 10cm from the table's center) into each pocket on the table, and verify the results by finding the correct event in the event log of the simulation. Our results are shown in table \ref{tab:experiment3D}. The pass rate shown is the percentage of successful trials from $\sim$1000 total trials of each method and model pairing. These results come to a very different outcome than the 2D case. Here, the addition of feedback from the simulation improves performance only marginally. We find that across LLMs and methods, there are very few successful trials, and these cases often contain flawed or unsatisfactory logic for the problem. We find that all the LLMs tested perform quite poorly at this type of task, independent of size. With larger LLMs, it has been shown that performance follows size for many metrics, but this hits a limit for complicated 3D reasoning task such as this, where no matter the quality of reasoning, no LLMs are currently capable of a reasonable pass rate. While still performing better than random selection, the LLMs are attempting the task but simply are not currently capable enough to succeed.

\section{Conclusion}
In this study, we have highlighted the poor performance of LLMs at parameter inference for physical reasoning. We  proposed the idea of exposing LLMs to simulation data to improve their performance. Our first attempt is to mimic human learning through experience (trial) and error, which appears to improve performance in 2D. Our method outperforms baseline CoT in 2D environments, and gains further improvements as the difficulty of the environment increases. However our work suggests that additional methods are needed for LLMs to effectively perform actions in simulated environments. \MethodName is a positive step towards effective reasoning within complex environments, but when pushed to a full 3D environment those improvements are lost, showing the upper limit of scene complexity that current LLMs can reasonably interact with.

We believe this study has presented evidence of a fundamental shortcoming of LLM capability. Although this outcome may be expected since they are trained on text alone, it is important to highlight such weaknesses considering their massive popularity. 
This could be useful information to approaches that seek to utilise LLMs in real/simulated environments because it shows that these models have poor/no intuition about physical reasoning.

A natural question from the reader might be ``why LLMs?''. While it is evident that several machine learning techniques have been developed particularly for parameter inference and system identification, the hope here is that the scalability of LLMs will create opportunities for further reasoning with natural language about the system. As these models evolve, it is foreseeable that they will offer more capability to reason about physics at a higher level. For example to answer a query such as: ``Is it possible to pot the 7-ball without displacing any other ball on the table?" with reference to the state visualized in Figure~\ref{pooltool}.
 
\bibliographystyle{unsrtnat}
\bibliography{references}

\end{document}